%% file: Draft2.tex
\documentclass[hidelinks]{siamart190516}
\usepackage{amsmath}
\usepackage{amssymb}
\usepackage{mathtools}
\usepackage{tikz}
\usepackage{pgfplots}
\usepackage{smalljeff}
\pgfplotsset{width=10cm, compat=1.9}
\usepgfplotslibrary{external,groupplots}
\usepgfplotslibrary{fillbetween}
\title{A Numerical Investigation of the Minimum Width of a Neural Network}
\author{Ibrohim Nosirov\thanks{
	SkyView Academy,
	6161 Business Center Drive,
	Highlands Ranch, CO 80130.
}\and
Mentor: Jeffrey M. Hokanson\thanks{
	Department of Computer Science, 
	University of Colorado Boulder,
	1111 Engineering Dr, Boulder, CO 80309,
	(\email{Jeffrey.Hokanson@colorado.edu}).}
}

\begin{document}
\maketitle
\begin{abstract}
Neural network width and depth are fundamental aspects of network topology. Universal approximation theorems provide that with increasing width or depth, there exists a neural network that approximates a function arbitrarily well. These theorems assume requirements, such as infinite data, that must be discretized in practice. Through numerical experiments, we seek to test the lower bounds established by Hanin in 2017.
\end{abstract}
\section{Introduction}
Neural networks are an increasingly popular choice in applications such as text recognition, computer vision, and modelling. 
In each of these applications, function approximation is a fundamental building block.
Neural networks are appealing due to their ability to approximate any continuous function with arbitrary accuracy.
For example, the Universal Approximation Theorem~\cite{Cybenko1989} states that a neural network with one hidden layer and increasing width is able to approximate any function with arbitrary accuracy.
However, most network topologies rarely consist of just a single hidden layer, with deep nets---networks with multiple hidden layers---being a more common use-case.
Hanin's universal approximation Theorem provides a similar result for these deep networks~\cite{Han17x}.
This theorem states that for fixed network widths (specifically $w \geq n+2$ for $n$ number of dimension) and increasing depth, a deep network can approximate any continuous function with arbitrary accuracy.
Here, we aim to numerically explore the utility of this result on functions of varying complexity.\\

\section{Background}
A neural network $\mathcal{N}$ is defined as a composition of affine transforms and a non-linear function. Given an input of dimension $n$, a net of depth $d$, width $w$, and non-linear function $g(\mathbf{x}) = \max{\{\mathbf{0}, \mathbf{x}\}}$ evaluated entry-wise, called rectified linear unit (ReLU), a neural network is
\begin{equation}
f_\mathcal{N}(x):= A_{d+1} \circ g \circ A_{d} \circ ...\circ g \circ A_0(x)
\end{equation} where $A_1, \ldots, A_d$ are affine transforms from $\mathbb{R}^w \to \mathbb{R}^w$, $A_0$ is an affine transform from $\mathbb{R}^n \to \mathbb{R}^w$ and $A_{d+1}$ is an affine transform from $\mathbb{R}^w \to \mathbb{R}^1$.
Here we seek to approximate the continuous function, letting  $f$ on the compact domain $D\subset \mathbb{R}^n$ by a neural network in the sup-norm:

\begin{equation}
\min_{\widetilde{f} \in \mathcal{N}_{w, d}} \|f - \widetilde{f} \|_\infty
\end{equation}
where the space of neural nets of width $w$ and depth $d$ is
\begin{equation}
\mathcal{\mathcal{N}}_{w, d} = \left\{
	f_\mathcal{N}(x): \parbox{25em}{is a fully connected neural net with properties defined in (1.1)}
	\right\}
\end{equation}
The Universal Approximation Theorem states that there exists a sequence of networks with increasing width $w$ and one hidden layer, $d = 1$, minimizing the difference between $f$ and $\widetilde{f}$.
\begin{equation}
\lim_{w \to \infty} \min_{\widetilde{f} \in \mathcal{N}_{w, 1}} \|f - \widetilde{f} \|_\infty = 0
\end{equation}

Hanin's theorem of universal approximation~\cite{Han17x} with bounded width states
\begin{equation}
\lim_{d \to \infty} \min_{\widetilde{f} \in \mathcal{N}_{n+2, d}} \|f - \widetilde{f} \|_\infty = 0\end{equation}

\section{Numerical Experiments}
To test Hanin's result of bounded width, we defined 3 functions from the virtual library of simulation experiments~\cite{SB13}.
We chose two dimensional function to densely populate the input space.
The Ackley function is a test problem with an oscillatory surface of cosine functions used in optimization,
\begin{equation}
	f_1(x, y) = -20 \exp\left(-0.2 \dfrac{1}{\sqrt{2}} \sqrt{x^2+y^2}\right) - \exp\left(\frac{1}{2}(\cos 2\pi x + \cos2\pi y) \right)+ 20 + e
\end{equation}
Rosenbrock is an optimization test problem with one global minimum in an uneven valley,
\begin{equation}
	f_2(x, y) = 100(x^4 - 2xy + y^2) + 1 -2x + x^2
\end{equation}
Borehole is an engineering test problem used to model fluid flow,
\begin{equation}
\begin{split}
	& f_3(r, L) = \frac{2\pi T_u (H_u - H_l)}{\ln(\frac{r}{r_w})\Big(1+\frac{2LT_u}{\ln(\frac{r}{r_w})r^2_w K_w}+\frac{T_u}{T_l}\Big)}\\
	& \text{where}\  r_w = 0.1,\  T_u =89335,\  H_u =1050,\  T_l = 89.55,\  H_l =760,\  K_w = 10950
\end{split}
\end{equation}

Since it is impossible to train on the full domain, we chose a discretized domain and sampled it using a grid.
The domain was sampled with a $320 \times 320 $ grid to create a training set. This number of points was chosen to balance computing time and sampling density. Grid sampling was used to ensure uniform distribution.
We used the TensorFlow package to model the data.
Each network was trained with the Adam algorithm, a modified version of stochastic gradient descent, and reinitialized with a random seed 200 times of which the lowest error was reported.
This was done to account for any local minimizers the network might fall into. 
The network was evaluated with a testing set on a $1000\times1000 $ grid on the domain.
A sequence of networks were created, trained and tested, gradually increasing the depth for every neural network width from 1 to width 10. For every width, 10 distinct neural networks (depths 1 through 10) were created to numerically test Hanin's result. Though Hanin's theorem proves existence, we explored whether networks of $w \geq n+2$ would behave differently than networks of smaller width.

\section{Results}
\input{fig_converge}
A monotonic decrease in error is expected as depth is increased, but due to local minimizers, the error does not decrease monotonically. 
According to Hanin, error should convergence toward zero in all networks with $w \geq n + 2$. 
Since our inputs are 2-dimensional, we should expect a significant drop in error for networks of width $w \geq 4$.
We see that many networks $w \geq n + 2$ , indeed experience a significant drop in error, the drop is not monotonic, with error momentarily increasing for some networks. 
However, nets with width $w \geq n+2$ had a visible difference in both convergence and overall error from networks of widths of $w = n - 1$ and $w = n$. The networks of width $w = n + 1$ behaved unpredictably, converging poorly in some instances, like in the Borehole function.\\
\newpage
\bibliographystyle{siamplain}
\bibliography{master}
\end{document}

%% file: fig_converge.tex
\definecolor{r1}{RGB}{158,1,66}

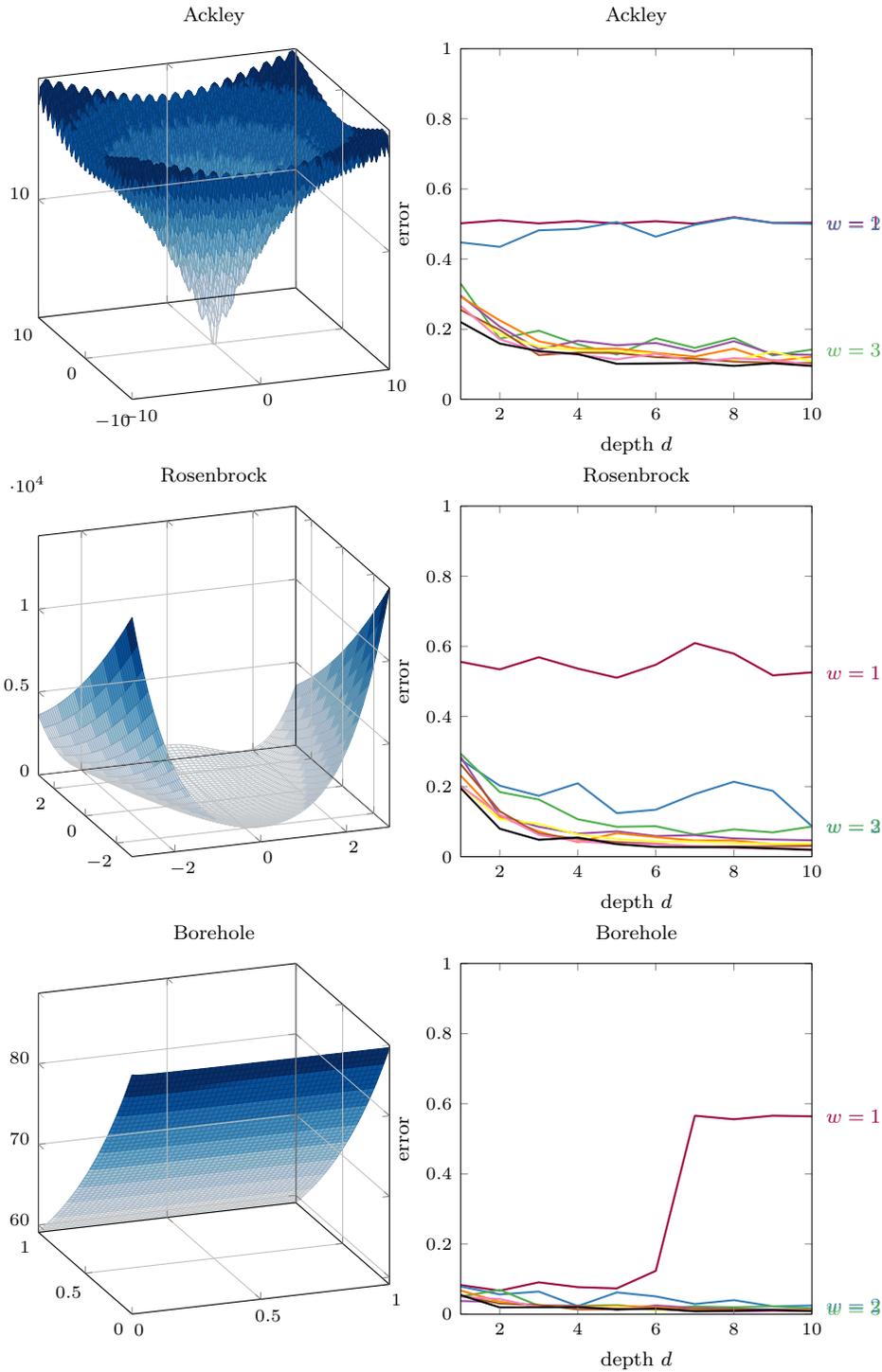
\begin{figure}
\begin{tikzpicture}
\begin{groupplot}[
	group style = {group size = 2 by 3, vertical sep = 1.5cm},
	width=0.5\linewidth,
	height=0.5\linewidth,
	]

	\nextgroupplot[view= {-20}{20}, grid=both, title = Ackley]
        \addplot3[surf] file {fig_ackley.dat};
	
	\nextgroupplot[
		xlabel = depth $d$,
		ylabel = error,
		title = Ackley,
		xmin = 1,
		xmax = 10,
		ymin = 0e0,
		ymax = 1,
		ymode = linear,
		clip = false,
	]
	\addplot[thick,r1] table [x=depth, y=width_1] {ackely.dat}
		node [pos=1, anchor=west, xshift=2pt] {$w=1$};
	\addplot[thick, colorbrewerA2] table [x=depth, y=width_2] {ackely.dat}
		node [pos=1, anchor=west, xshift=2pt] {$w=2$};
	\addplot[thick, colorbrewerA3] table [x=depth, y=width_3] {ackely.dat}
		node [pos=1, anchor=west, xshift=2pt] {$w=3$};
	\addplot[thick, colorbrewerA4] table [x=depth, y=width_4] {ackely.dat};
	\addplot[thick, colorbrewerA5] table [x=depth, y=width_5] {ackely.dat};
	\addplot[thick, colorbrewerA6] table [x=depth, y=width_6] {ackely.dat};
	\addplot[thick, colorbrewerA7] table [x=depth, y=width_7] {ackely.dat};
	\addplot[thick, colorbrewerA8] table [x=depth, y=width_8] {ackely.dat};
	\addplot[thick] table [x=depth, y=width_9] {ackely.dat};

	\nextgroupplot[view= {-20}{20}, grid=both, title = Rosenbrock]
        \addplot3[surf] file {fig_rosenbrock.dat};

	\nextgroupplot[
		xlabel = depth $d$,
		ylabel = error,
		title = Rosenbrock,
		xmin = 1,
		xmax = 10,
		ymin = 0e0,
		ymax = 1,
		ymode = linear,
		clip = false,
	]
	\addplot[thick,r1] table [x=depth, y=width_1] {rosenbrock.dat}
		node [pos=1, anchor=west, xshift=2pt] {$w=1$};
	\addplot[thick, colorbrewerA2] table [x=depth, y=width_2] {rosenbrock.dat}
		node [pos=1, anchor=west, xshift=2pt] {$w=2$};
	\addplot[thick, colorbrewerA3] table [x=depth, y=width_3] {rosenbrock.dat}
		node [pos=1, anchor=west, xshift=2pt] {$w=3$};
	\addplot[thick, colorbrewerA4] table [x=depth, y=width_4] {rosenbrock.dat};
	\addplot[thick, colorbrewerA5] table [x=depth, y=width_5] {rosenbrock.dat};
	\addplot[thick, colorbrewerA6] table [x=depth, y=width_6] {rosenbrock.dat};
	\addplot[thick, colorbrewerA7] table [x=depth, y=width_7] {rosenbrock.dat};
	\addplot[thick, colorbrewerA8] table [x=depth, y=width_8] {rosenbrock.dat};
	\addplot[thick] table [x=depth, y=width_9] {rosenbrock.dat};
	
	\nextgroupplot[view= {-20}{20}, grid=both, title = Borehole]
        \addplot3[surf] file {fig_borehole.dat};
	
	\nextgroupplot[
		xlabel = depth $d$,
		ylabel = error,
		title = Borehole,
		xmin = 1,
		xmax = 10,
		ymin = 0e0,
		ymax = 1,
		ymode = linear,
		clip = false,
	]
	\addplot[thick,r1] table [x=depth, y=width_1] {borehole.dat}
		node [pos=1, anchor=west, xshift=2pt] {$w=1$};
	\addplot[thick, colorbrewerA2] table [x=depth, y=width_2] {borehole.dat}
		node [pos=1, anchor=west, xshift=2pt] {$w=2$};
	\addplot[thick, colorbrewerA3] table [x=depth, y=width_3] {borehole.dat}
		node [pos=1, anchor=west, xshift=2pt] {$w=3$};
	\addplot[thick, colorbrewerA4] table [x=depth, y=width_4] {borehole.dat};
	\addplot[thick, colorbrewerA5] table [x=depth, y=width_5] {borehole.dat};
	\addplot[thick, colorbrewerA6] table [x=depth, y=width_6] {borehole.dat};
	\addplot[thick, colorbrewerA7] table [x=depth, y=width_7] {borehole.dat};
	\addplot[thick, colorbrewerA8] table [x=depth, y=width_8] {borehole.dat};
	\addplot[thick] table [x=depth, y=width_9] {borehole.dat};

\end{groupplot}
\end{tikzpicture}
\caption{The result of the three numerical experiments, which measured error in approximation defined as $\|f - \widetilde{f} \|_\infty \geq \max \|f(x_j)-\widetilde{f}(x_j)\|.$ On the left we see the function we are approximating; the right plot shows the error.}

\end{figure}